\documentclass[letterpaper, 10 pt, conference]{ieeeconf}  

\IEEEoverridecommandlockouts                        
\usepackage{xcolor}
\definecolor{wong-black}        {HTML}{000000}
\definecolor{wong-lightorange}  {HTML}{E69F00}
\definecolor{wong-lightblue}    {HTML}{56B4E9}
\definecolor{wong-green}        {HTML}{009E73}
\definecolor{wong-yellow}       {HTML}{F0E442}
\definecolor{wong-darkblue}     {HTML}{0072B2}
\definecolor{wong-darkorange}   {HTML}{D55E00}
\definecolor{wong-pink}         {HTML}{CC79A7}

\usepackage[accsupp]{axessibility}  

\usepackage{url}

\usepackage{hyperref} 

\hypersetup{
    colorlinks=true,
    citecolor=wong-green,
    linkcolor=wong-darkblue,
    filecolor=wong-pink,      
    urlcolor=wong-black
    }

\usepackage{amsthm}
\usepackage[T1]{fontenc}
\usepackage{cite}
\usepackage{amsmath,amssymb,amsfonts}
\usepackage{algorithmic}
\usepackage[pdftex]{graphicx}
\usepackage{booktabs}
\usepackage{textcomp}
\usepackage{svg}
\usepackage[nolist, nohyperlinks, printonlyused]{acronym} 
\usepackage{cuted}

\usepackage{caption}
\usepackage{subcaption}
\usepackage{tikz}
\usepackage{etoolbox}
\usepackage{color}
\usetikzlibrary{shapes.geometric, arrows, positioning}
\usepackage{verbatim} 
\usepackage{xargs}
\usepackage{mathtools}
\usepackage{textcomp}
\usepackage{xcolor}
\def\BibTeX{{\rm B\kern-.05em{\sc i\kern-.025em b}\kern-.08em
    T\kern-.1667em\lower.7ex\hbox{E}\kern-.125emX}}

\theoremstyle{definition}

\def\BibTeX{{\rm B\kern-.05em{\sc i\kern-.025em b}\kern-.08em
    T\kern-.1667em\lower.7ex\hbox{E}\kern-.125emX}}

\usepackage{eso-pic}
\newcommand{\mycopyrighttext}{%
  \footnotesize
  \noindent
  \textcopyright~2025 IEEE. Personal use of this material is permitted. Permission from IEEE must be obtained for all other uses, in any current or future media, including reprinting/republishing this material for advertising or promotional purposes, creating new collective works, for resale or redistribution to servers or lists, or reuse of any copyrighted component of this work in other works.\\
  IEEE 36th Intelligent Vehicles Symposium (IV 2025) - 22-25 June, 2025.
}

\AtBeginDocument{%
  \AddToShipoutPicture*{%
    \AtPageUpperLeft{%
      \put(55, -40){
        \parbox{\dimexpr\textwidth-4pt\relax}{\raggedright \mycopyrighttext}
      }%
    }
  }
}

\begin{document}


\title{\LARGE \bf
Balancing Progress and Safety: A Novel Risk-Aware \\Objective for RL in Autonomous Driving
}

\author{Ahmed Abouelazm$^{1}$, Jonas Michel$^{1,2}$, Helen Gremmelmaier$^{1}$, \\ Tim Joseph$^{1}$, Philip Schörner$^{1}$, and J. Marius Zöllner$^{1,2}$
\thanks{$^{1}$Authors are with the FZI Research Center for Information Technology, Germany
        {\tt\small abouelazm@fzi.de}}%
\thanks{$^{2}$Authors are with the Karlsruhe Institute of Technology, Germany}%
}

\maketitle


\begin{acronym}
    \acro{ml}[ML]{Machine Learning}
    \acro{cnn}[CNN]{Convolutional Neural Network}
    \acro{dl}[DL]{Deep Learning}
    \acro{ad}[AD]{Autonomous Driving}
\end{acronym}


\begin{abstract}
    Reinforcement Learning (RL) is a promising approach for achieving autonomous driving due to robust decision-making capabilities. RL learns a driving policy through trial and error in traffic scenarios, guided by a reward function that combines the driving objectives. The design of such reward function has received insufficient attention, yielding ill-defined rewards with various pitfalls. Safety, in particular, has long been regarded only as a penalty for collisions. This leaves the risks associated with actions leading up to a collision unaddressed, limiting the applicability of RL in real-world scenarios. To address these shortcomings, our work focuses on enhancing the reward formulation by defining a set of driving objectives and structuring them hierarchically. Furthermore, we discuss the formulation of these objectives in a normalized manner to transparently determine their contribution to the overall reward. Additionally, we introduce a novel risk-aware objective for various driving interactions based on a two-dimensional ellipsoid function and an extension of Responsibility-Sensitive Safety (RSS) concepts. We evaluate the efficacy of our proposed reward in unsignalized intersection scenarios with varying traffic densities. The approach decreases collision rates by 21\% on average compared to baseline rewards and consistently surpasses them in route progress and cumulative reward, demonstrating its capability to promote safer driving behaviors while maintaining high-performance levels.
    
       
\end{abstract}



\section{Introduction}
\label{sec:introduction}
End-to-End (E2E) approaches for autonomous driving have gained significant attention in recent years~\cite{endtoendreview}. Unlike modular frameworks, which divide the driving task into distinct components such as perception, prediction, and planning — each requiring significant manual design, E2E approaches directly learn decision-making from raw sensor inputs. This enables neural networks to learn efficient representations of their interactions with the environment, minimizing reliance on human-engineered modules. As a result, they achieve greater flexibility and adaptability by dynamically learning from real-world experiences~\cite{endtoendreviewpaper}.

E2E driving can be realized through Reinforcement Learning (RL). RL offers a flexible and adaptable learning framework where agents learn through direct interactions with the environment~\cite{Sutton1998}. The reward function guides the learning process, reinforcing actions that yield desirable outcomes and penalizing ineffective and dangerous behaviors~\cite{kiran2021deep}. By maximizing cumulative reward, RL agents improve their decision-making, enabling them to navigate complex environments and achieve their goals~\cite{deeprlintro}.

\textbf{Research Gap.}
Autonomous driving is a complex domain characterized by multiple objectives, often conflicting in nature, such as balancing progress toward the goal with ensuring safety. Designing a reward function that effectively captures these complexities is a significant challenge~\cite{misdesign}. Inadequately designed reward functions can result in suboptimal driving policies and potentially hazardous behaviors, particularly in critical situations. Despite their importance, recent RL research has often underemphasized reward design, focusing instead on learning algorithms or handling different state and action spaces~\cite{ahmed_reward_review}. In this paper, we aim to address this oversight by improving the reward function to better capture the complexities of autonomous driving.
\begin{figure}[t!]
    \begin{center}
    \includegraphics[width=0.85\columnwidth, ]{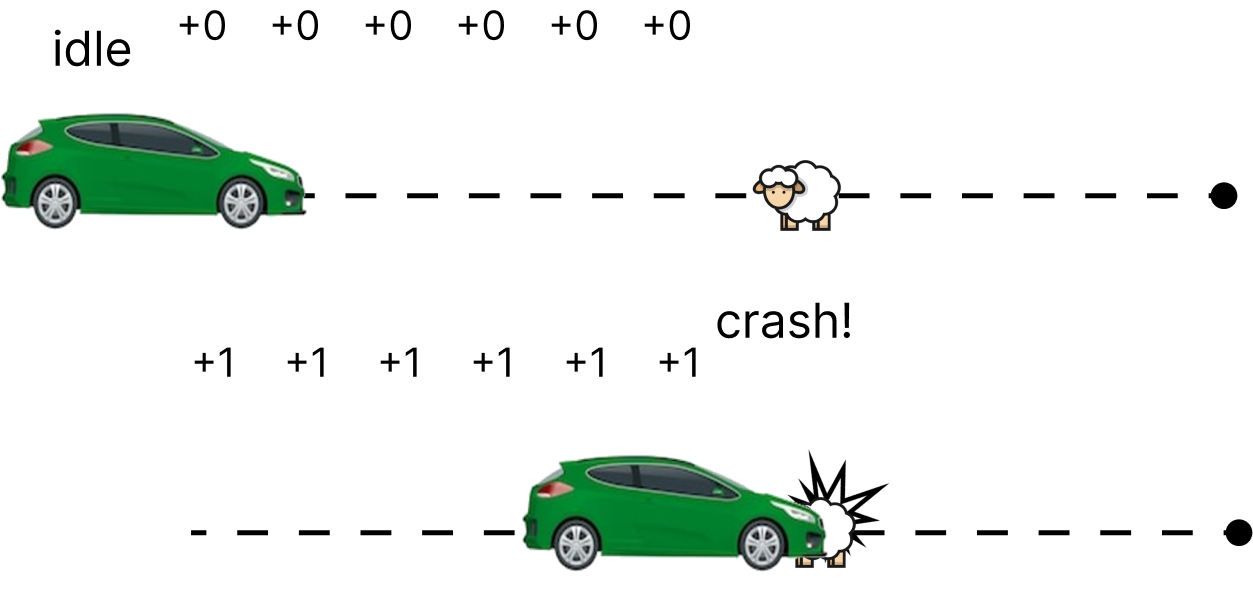}
    \caption{Example of irrational behavior in RL agents caused by sparse safety objective formulation and poor handling of the conflict between safety and progress objectives~\cite{misdesign}.}
    \label{fig:reward mis-design example}
    \end{center}
    \vspace{-0.6cm} 
\end{figure}


This research is particularly driven by insights from~\cite{misdesign}, which identifies different failure cases arising from poorly designed rewards commonly employed in autonomous driving. Fig.~\ref{fig:reward mis-design example} demonstrates a common misdesign in the reward function that leads to catastrophic outcomes, even in a simple scenario where a single static obstacle blocks the agent's path to its goal. Unlike a human driver, who would wait indefinitely in such a scenario while the obstacle persists, the agent can opt to collide with the obstacle rather than remain stationary. This seemingly irrational behavior arises because the accumulated progress penalty associated with extended waiting can outweigh the collision penalty. This insufficient handling of the conflict between safety and progress stems from the sparsity of the safety objective, only penalizing collision without considering the associated risk of preceding actions. Adjusting the weights assigned to the safety and progress objectives may mitigate this issue. Still, it fails to resolve the underlying problem fully.

\textbf{Contribution.} This work enhances the driving policies of RL agents by analyzing and extending the formulation of reward objectives, with a particular emphasis on risk awareness. The key contributions of this work are:

\begin{itemize}
    \item \textbf{Hierarchical Structuring of Objectives}: Introducing a structured reward function that organizes driving objectives hierarchically as a directed graph.
    \item \textbf{Enhanced Objective Formulation}: Proposing refined and normalized formulations for each driving objective to improve interoperability and comparability.
    \item \textbf{Advanced Risk-Aware Objective}: Developing a driving risk objective that integrates geometric and dynamic risk factors through a two-dimensional ellipsoid model.
\end{itemize}

\section{Related Work}
\label{sec:related_work}
In this section, we review related work on the use of RL in autonomous driving, covering different decision-making levels, sensor configurations, and action spaces, as well as recent advancements in reward function design.

\subsection{RL for Autonomous Driving}
RL has been utilized in autonomous driving at various decision-making levels. In works such as~\cite{caoHighwayExitingPlanner2021}, 
RL is applied as a high-level behavioral planner for lane-assistance systems and turning maneuvers. 
In such a setting, RL impacts only the agent's behavior, and an additional trajectory planner is required to align agent motion with desired behavior~\cite{behavioralvstrajectory}. Moreover, RL can be employed for trajectory planning, where the agent estimates a sequence of waypoints over a planning horizon. The task of trajectory planning has been tackled in Cartesian frame~\cite{naveed2021trajectory} 
as well as Frenet space~\cite{bogdoll2024informed}. 
Finally, RL can directly estimate control inputs such as steering and acceleration, eliminating the necessity for additional components~\cite{wu2022uncertainty}.

In addition to spanning several decision-making levels, RL supports a diverse range of input modalities. In multiple works, object-level information serves as an input observation, with objects typically described by their location, velocity, and acceleration in either a global frame or relative to the self-driving vehicle
~\cite{naveed2021trajectory}. 
Alternative papers instead learn driving policies directly from raw sensor data. Multiple approaches utilize either camera images~\cite{bogdoll2024informed} 
or LiDAR sensors~\cite{mammadov2023end} to perceive the current environment state. Recent advancements in RL incorporate multiple sensor modalities to realize the driving task~\cite{xiao2020multimodal}. 

\subsection{Reward Functions for Autonomous Driving}
Driving rewards should encapsulate the conflicting objectives of the driving task into a single numerical value to guide RL agent decision-making. Building on the review established in~\cite{ahmed_reward_review}, this research breaks down driving rewards into key objectives such as safety, progress, comfort, and conformance to traffic rules.

The safety objective encourages the agent to avoid collision with other road users and static obstacles. This objective is commonly a sparse penalty at collision~\cite{chen2019model}. Only a few studies attempted to improve the safety objective by introducing a penalty for near-misses~\cite{r11}, as well as simplified calculations of time to collision (TTC)~\cite{r20} and headway~\cite{li2018urban}. These studies have shown an enhancement in the overall safety of the agents. Recognizing the criticality of the safety objective, our work argues that the simplified formulations in previous studies fail to fully capture the complexities of real-world driving environments, highlighting significant potential for further improvement. To address this, this paper proposes a risk-aware objective that builds upon conventional safety measures by drawing on insights from recent advances in safety frameworks, such as the Nvidia Force Field (SFF) \cite{nvidiaforcefield} and Responsibility-Sensitive Safety (RSS) \cite{rss}.

The progress objective incentivizes the agent to advance towards its goal~\cite{r20}, while comfort promotes smooth and comfortable trajectories during navigation~\cite{r7}. Lastly, traffic conformance assesses the agent's adherence to traffic regulations, primarily focusing on maintaining the speed limit~\cite{chen2019model} and staying within the lane~\cite{chen2021interpretable}. Further limitations and a discussion of the reward function and its components are elaborated in~\cite{misdesign,ahmed_reward_review}. Alternatively, it is feasible to learn a reward function from expert data using Inverse Reinforcement Learning (IRL) as proposed in~\cite{zhao2022personalized}.
Despite its potential for reward learning, IRL demands a large amount of high-quality data, exhibits limited generalization, and lacks explainability. This may result in a reward function that shapes unexpected behaviors in RL policy. 

Another aspect to consider regarding reward function design is handling conflicting objectives and prioritizing them effectively, as illustrated in the case study presented in Section \ref{sec:introduction}. Most prior research assigns weights to each objective based on expert knowledge~\cite{zhu2020safe,chen2021interpretable}. However, this method requires manual adjustment and adaptation for different scenarios, which is a clear drawback. Alternatively, a more fitting approach to the problem's nature is to treat the reward as a multi-objective problem with a hierarchical order between objectives~\cite{rulebook}. 
This work explores a suitable reward structure and various methods to consolidate its objectives into a unified value.

\section{Method}
\label{sec:method}
This section outlines our methodology for improving reward functions in autonomous driving, which involves breaking down the reward into a set of distinct objectives and establishing a hierarchical structure to assign weights to these objectives. Furthermore, we introduce formulations for each objective alongside a novel risk awareness objective.
\subsection{Reward Structure and Weights Assignment}
\begin{figure}[t!]
    \centering
\includegraphics[width=0.9\columnwidth,keepaspectratio]{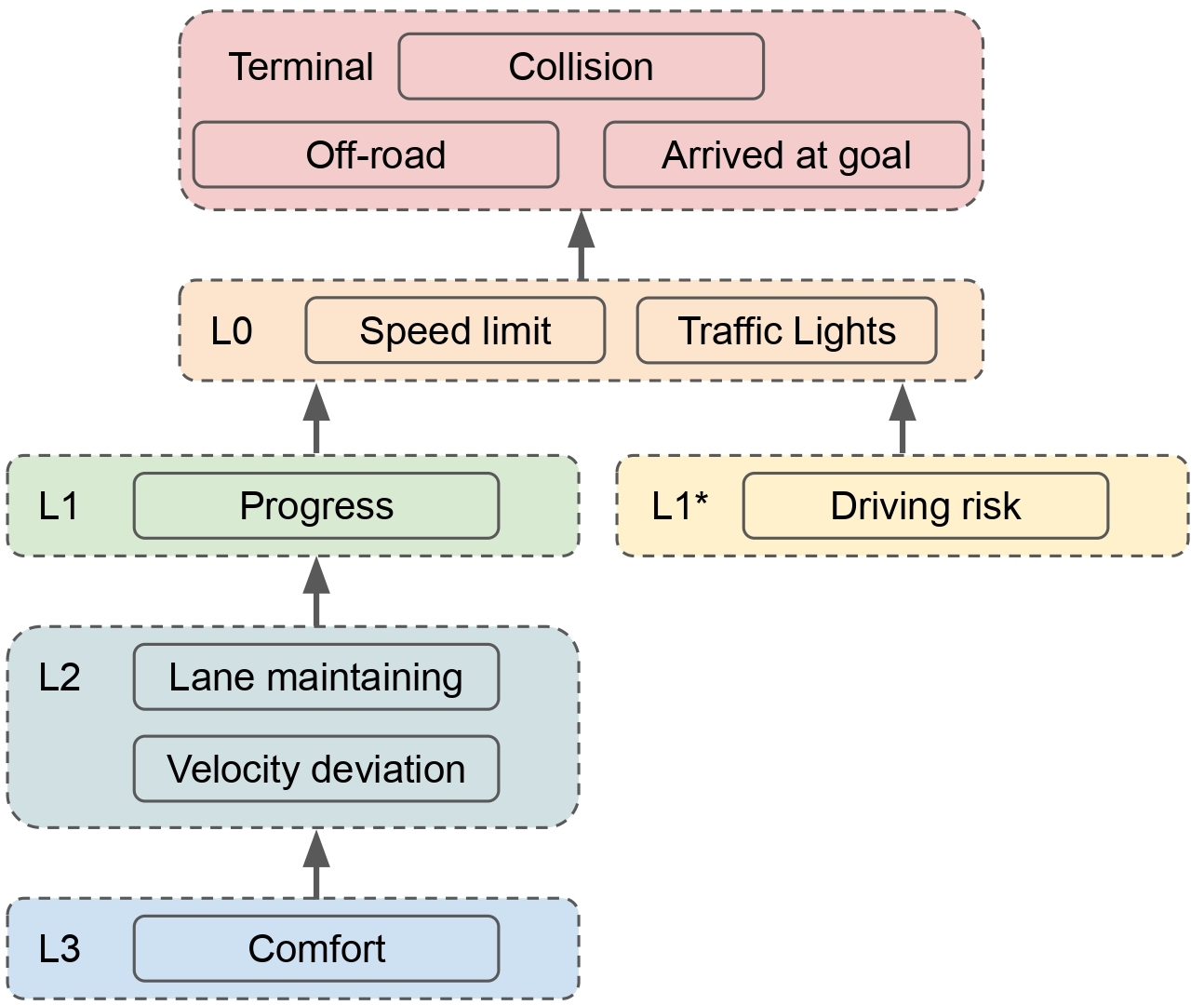}
    \caption{Representation of the driving reward as a directed graph, where the objective level indicates its priority.}
    \label{fig:rulebook}
    \vspace{-0.6cm} 
\end{figure}
We decompose the driving reward into four core objectives, safety, progress, comfort, and traffic rule conformance, guided by insights from~\cite{ahmed_reward_review}. To represent dependencies and potential conflicts among objectives, as discussed in section \ref{sec:introduction}, we arrange the objectives into a priority-based directed graph inspired by Rulebooks~\cite{rulebook}. A Rulebook organizes rules hierarchically, as a directed graph~\cite{collin2020safety}, enabling actions to be evaluated in a top-down manner. Each rule is expressed as a scalar value indicating its degree of violation.

Our reward structure organizes each objective hierarchically at a specific level \(L\) as a set of terms, as shown in Fig.~\ref{fig:rulebook}. At the highest priority, the structure includes conditions that define the termination of a driving scenario, such as collisions with vehicles or static obstacles, leaving the permissible driving area, or reaching the destination. Level \(L_0\) addresses traffic rule conformance, enforcing specific rules such as adhering to speed limits and stopping at red lights. Level \(L_1\) encodes the progress objective, measured by the distance traveled towards the destination, serving as the primary motivation for the agent to advance. Equally important, Level \(L_1^*\) represents driving risk, a factor often completely disregarded or oversimplified in prior works using metrics like TTC or headway. This paper highlights the significance of driving risk and introduces a comprehensive formulation to better inform the agent about the quality of its actions.

Level \(L_2\) refines the progress objective with two terms: maintaining proximity to the centerline and minimizing deviation from the desired velocity. Level \(L_3\) addresses comfort and smoothness, promoting a better driving style as the agent progresses toward its destination. While Rulebook-based approaches often assign the progress objective \(L_1\) the lowest priority~\cite{collin2020safety}, this is not the case in our work. In reinforcement learning, objectives are combined into a weighted sum, where assigning sufficient weight to progress and shaping it with other terms is both common and effective. Assigning progress to the lowest weight risks the agent converging to a policy of always remaining stationary. Furthermore, we establish a normalized reward for each objective, ranging from zero to one. Normalization ensures all objectives share the same scale, making assigning priority weights more transparent and straightforward.

During the reward evaluation, if one of the terminal conditions is satisfied, only that specific condition contributes to the reward value. Otherwise, the reward is calculated as a weighted sum of rewards at each level, \(\mathcal{R}_{L_i}\), weighted by \(w_{L_i}\), as shown in Eq.~\ref{eq:rulebook_rewards}. The weights \(w_{L_i}\) are determined by the position of each level in the hierarchy, with higher-priority objectives assigned greater weights, following the methodology in~\cite{bogdoll2024informed}. Each weight \(w_{L_i}\) is computed using a base weight \(\beta\) and the level \(L_i\), as defined in Eq.~\ref{eq:rulebook_weights}.
\begin{equation}
    \mathcal{R}_{\,rulebook} = 
    \begin{cases} 
    \mathcal{R}_{Terminal} &, \text{if } \text{Terminal}\\ 
    \sum_{i=1}^{3} w_{L_i} \cdot \mathcal{R}_{L_i} + \mathcal{R}_{L_0}& ,\text{otherwise} 
    \end{cases}
    \label{eq:rulebook_rewards}
\end{equation}
\begin{equation}
    w_{L_i} = \beta ^{\,i-1}, \quad \text{where } \beta < 1
    \label{eq:rulebook_weights}
\end{equation}
\subsection{Safe and Risk-aware Reward Design}
In this section, we outline the objectives of each level and their corresponding rewards, with a particular emphasis on the enhancements in safety and risk awareness.
\subsubsection*{\textbf{Terminal Conditions}}
This level represents the terminal conditions of a driving scenario and the reward associated with these conditions. All terminal rewards are scaled by a weight \(w_{\,\text{Terminal}}\), which was empirically tuned, and the terminal reward is calculated as shown in Eq.~\ref{eq:terminal}.
\begin{equation}
    {\mathcal{R}}_{\,Terminal} = 
    w_{\,Terminal}\begin{cases}
        {\mathcal{R}}_{\,success}&,\text{If at goal}\\ 
        {\mathcal{R}}_{\,collision}&,\text{If collision}\\ {\mathcal{R}}_{\,offroad}&,\text{If off-road}
    \end{cases}
    \label{eq:terminal}
\end{equation}
The RL agent receives a penalty on collision with other agents or static obstacles in the environment, as depicted in Eq.~\ref{eq:r_s_t}. We incorporate a modified version of the collision penalty inspired by~\cite{r6}. This modified version integrates an adaptive term that depends on the ratio between the agent's velocity $v$ at the time of the collision and its maximum attainable velocity $v_{max}$ to differentiate collision severity. This version encourages the agent to slow down in anticipation of a potential collision and, if a collision occurs, to reduce its velocity to the lowest possible level.
\begin{equation}
    {\mathcal{R}}_{\,collision} = -1 \cdot \left (0.5 + 0.5 \cdot\,\dfrac{v}{v_{max}}\right )
    \label{eq:r_s_t}
\end{equation}
Additionally, the agent is required to stay within the permissible driving area and avoid veering off-road. A violation results in a strict penalty \({\mathcal{R}}_{\,\text{offroad}} = -1\), and immediate termination of the episode.
A reward is assigned to the agent upon successfully reaching its goal without violating the other two terminal conditions. We improve this success reward by incorporating the lateral offset at the goal. Specifically, if the agent's lateral offset upon reaching the goal is below a specified threshold, it earns a higher reward compared to being farther away, as defined in Eq.~\ref{eq:success_progress}.
\begin{equation}
    {\mathcal{R}}_{\,success} = 
    \begin{cases}
        1 & , \text{If } {\text{offset}}_{\,t} \,<  {\text{threshold}}_{\,\text{offset}} \\ 
        0.5 & , \text{otherwise}
    \end{cases}
    \label{eq:success_progress}
\end{equation}
Additionally, this work avoids penalizing timeouts, which occur when the agent exceeds the maximum number of steps without completing a driving task. Instead, the episode is terminated without any penalties. This approach prevents discouraging extended waiting periods, which may sometimes be necessary, such as when waiting for other vehicles to pass at an intersection.
\subsubsection*{\textbf{Traffic Rule Conformance (\(L_0\))}}This objective is designed to encourage the agent to adhere to traffic regulations and social norms, such as obeying speed limits and stopping at red lights. Traffic rules are implemented as a soft constraint, meaning violations incur a penalty, as shown in Eq.~\ref{eq:traffic_rules}, but do not terminate the episode. 
More complex traffic rules and social norms, such as yielding to pedestrians at crosswalks, are not represented in this work, as their formulation and integration into autonomous systems have been explored in other work~\cite{rizaldi2015formalising}.
\begin{equation}
    {\mathcal{R}}_{L_0} = 
    \begin{cases}
        -1 & , \text{If a traffic rule is violated}\\ 
        0 & , \text{otherwise}
    \end{cases}
    \label{eq:traffic_rules}
\end{equation}
\subsubsection*{\textbf{Safety and Risk Awareness (\(L_1^*\))}}
Enhancing safety through risk awareness objective is the cornerstone of our proposed reward function. We aim to construct a driving risk objective that improves the agent's risk awareness regarding its interaction with static obstacles and surrounding vehicles. This proposed driving risk accounts for both the geometry of interacting traffic participants and their dynamic behavior. While prior studies have relied on metrics such as TTC~\cite{r20}, and headway~\cite{li2018urban}, we focus on utilizing a more complex safety concept, namely RSS~\cite{rss}, as it offers a framework that surpasses the metrics previously employed. 

\begin{figure}[t]
\centering
\begin{minipage}[t]{0.5\columnwidth}
    \begin{subfigure}[t]{0.75\columnwidth}
        \centering
        \includegraphics[width=\columnwidth]{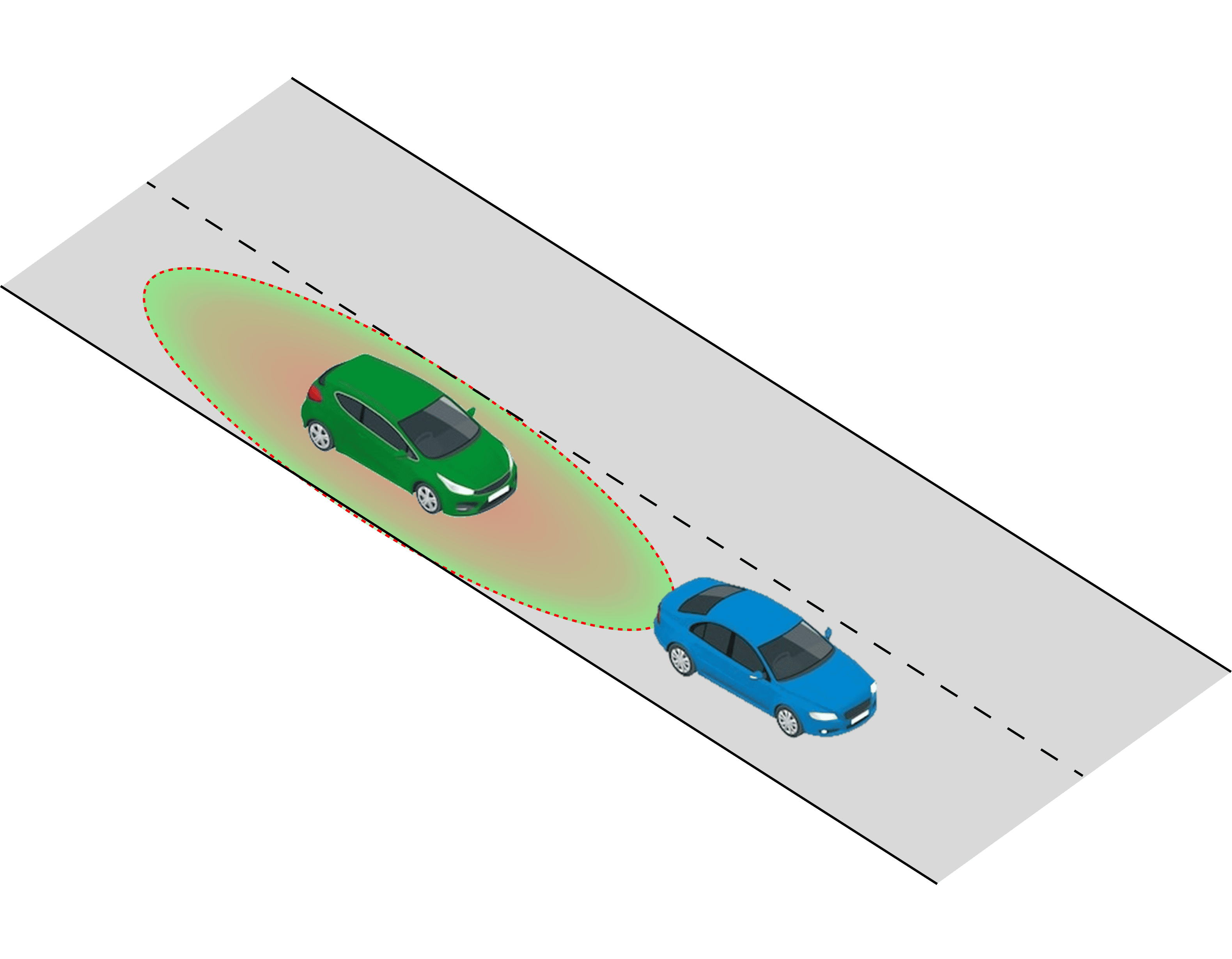}
        \caption{Interaction between two vehicles traveling in the same direction.}
        \label{fig:long_same}
    \end{subfigure}
    \vfill
    \vspace{2pt}
    \begin{subfigure}[t]{0.75\columnwidth}
        \centering
        \includegraphics[width=\columnwidth]{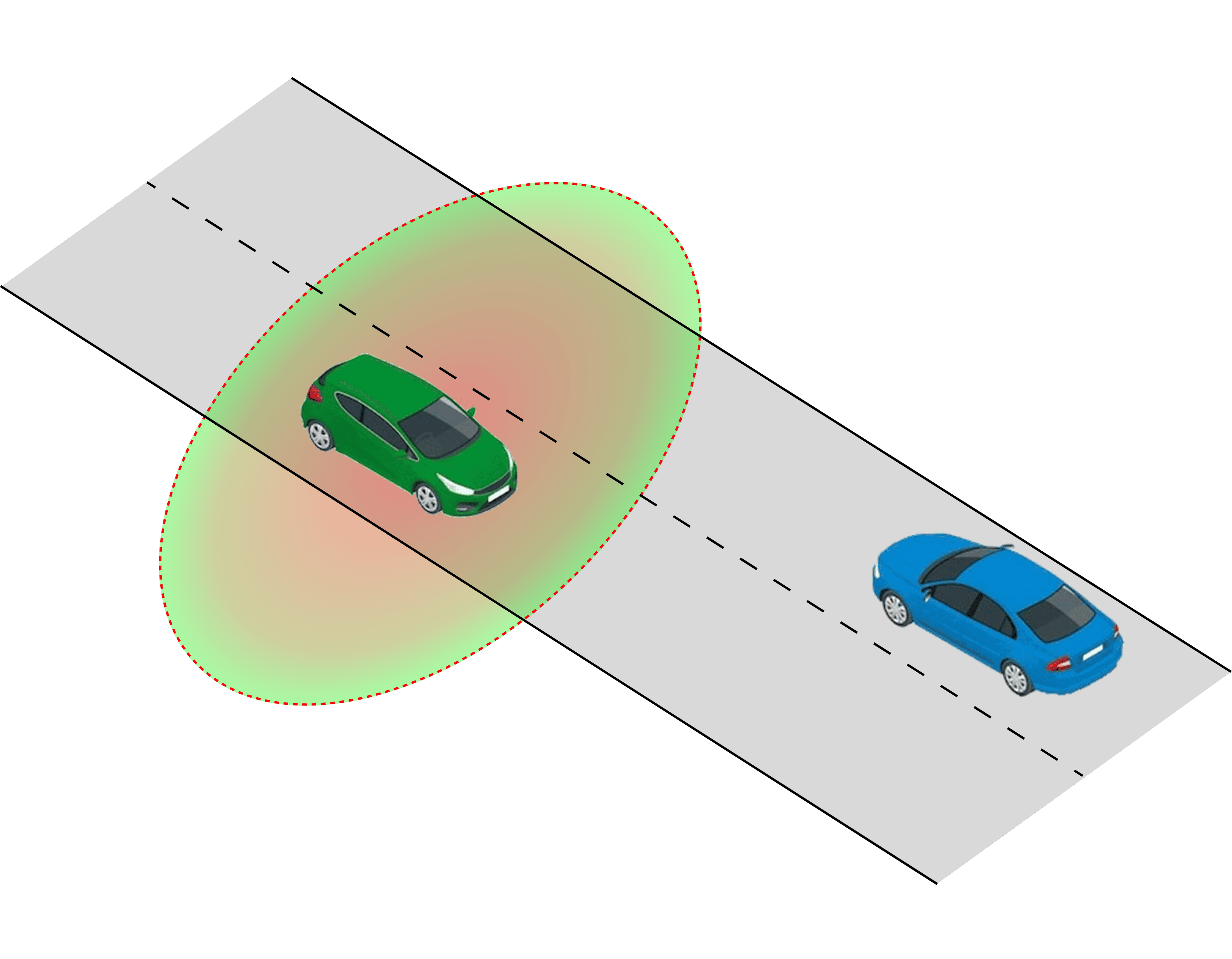}
        \caption{Interaction between two vehicles traveling in the opposite direction.}
        \label{fig:long_opp}
    \end{subfigure}
\end{minipage}%
\begin{minipage}[t]{0.5\columnwidth}
    \begin{subfigure}[t]{0.75\columnwidth}
        \centering
        \includegraphics[width=\columnwidth]{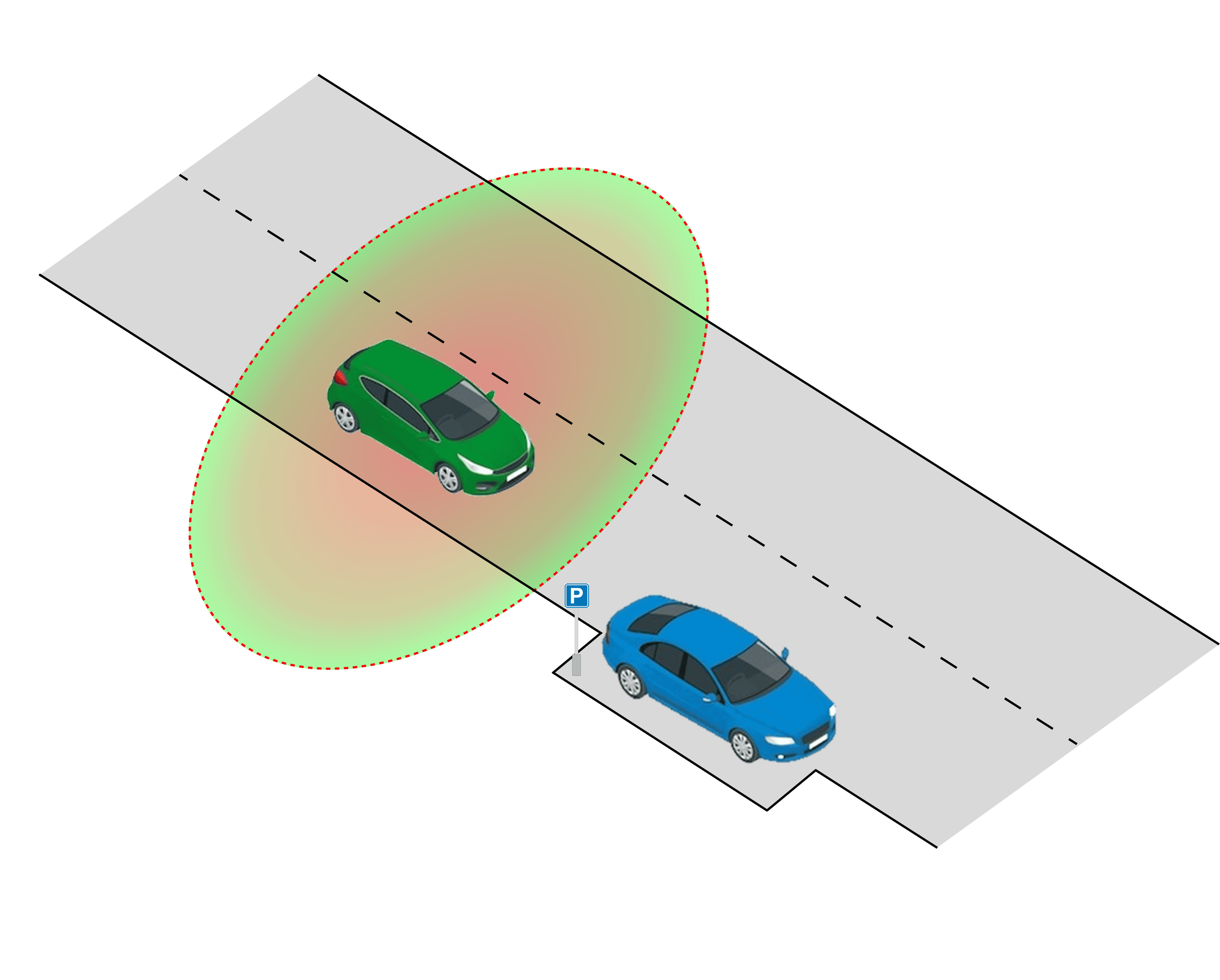}
        \caption{Interaction between a vehicle and a static obstacle.}
        \label{fig:long_static}
    \end{subfigure}
    \vfill
    \begin{subfigure}[t]{0.88\columnwidth}
        \centering
        \includegraphics[width=\columnwidth]{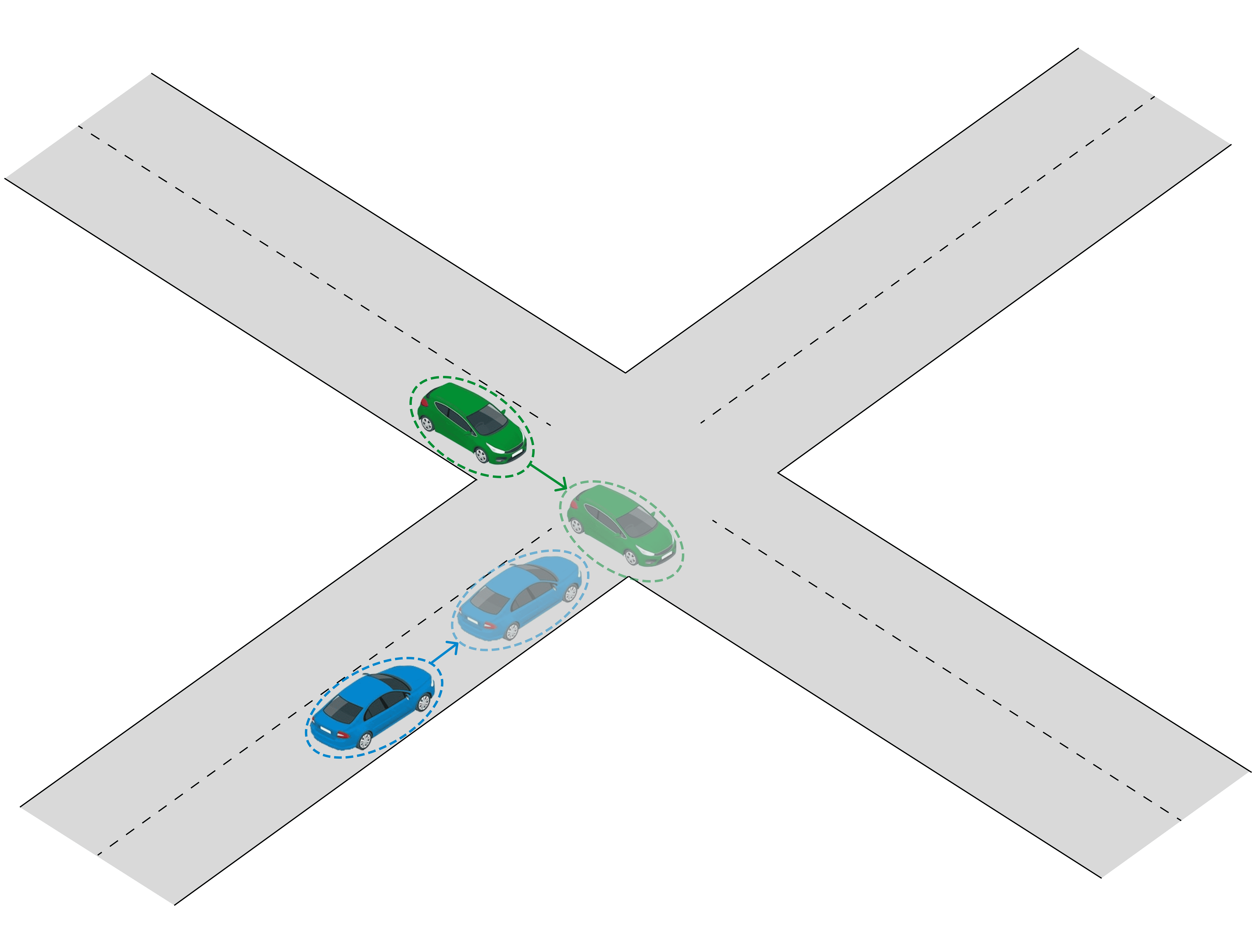}
        \caption{Interaction between two intersecting vehicles.}
    \end{subfigure}
\end{minipage}
\caption{The risk-aware objective considers different interactions between traffic participants. The RL agent is highlighted in green, and Non-Player Characters (NPCs) and obstacles are in blue.}
\label{fig:interaction_types}
\vspace{-0.7cm} 
\end{figure}

RSS~\cite{rss} is originally developed as a safety framework rather than a risk estimation. It aims to determine the minimum safety clearance between interacting traffic participants in both longitudinal and lateral directions using worst-case analysis. Notably, RSS has been validated through real-world implementations in advanced driver assistance systems and pilot autonomous driving projects, demonstrating its practical efficacy. This work extends and adapts the RSS framework to design a risk-aware objective for three key interaction modes between vehicles: same direction, opposite direction, and intersecting, as well as interactions involving static obstacles. These interaction modes collectively encompass all possible interactions encountered in driving scenarios. While Fig.~\ref{fig:interaction_types} depicts these interaction modes in single-lane scenarios, the proposed approach is directly applicable to multi-lane environments without any additional modifications.

The proposed risk-aware objective calculates two types of safety clearances: one based on the geometries of the interacting traffic participants and the other on their dynamic behaviors. Both clearances are converted into penalties $\left({\mathcal{P}}_{\,risk}^{\,geom}, {\mathcal{P}}_{\,risk}^{\,dyn}\right)$ using a non-linear ellipsoid function~\cite{typaldos2022optimization,karalakou2023deep}, referred to as a "risk field". While the ellipsoid function is unified, described in Eq.~\ref{eq:risk_field}, it employs distinct parameters tailored to each type of clearance. This flexibility allows the function to effectively assess combined penalties for both longitudinal ($x$) and lateral ($y$) directions. Penalties are computed based on the distances ($d_x$, $d_y$) between interacting agents and the parameters defining the ellipse shape, ensuring adaptability to different safety considerations.
\begin{equation}
    {\mathcal{P}}_{\,risk}= \left [ \left ( \dfrac{\left | d_x \right | - c_x}{r_x}\right)^{P_x}  +  \left ( \dfrac{\left | d_y \right | - c_y}{r_y}\right)^{P_y} + 1 \right ]^{-P}
    \label{eq:risk_field}
\end{equation}

\begin{figure}[t]
\centering
\includegraphics[width=0.9\columnwidth]{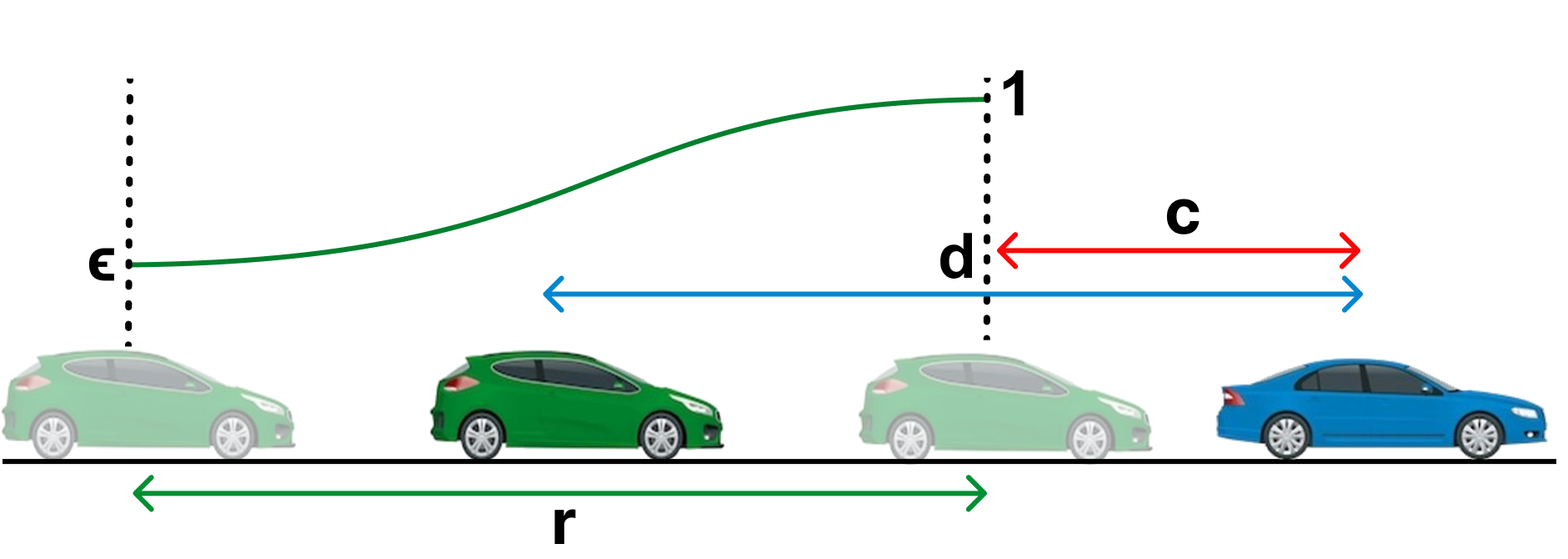}
    \caption{Visualization of one-dimensional ellipsoid penalty based on the distance between RL agent (green) and another vehicle (blue). The risk field ranges from 1 to $\epsilon$, where $\epsilon$ is close to zero.}
    \label{fig:risk_field_1d}
\vspace{-0.5cm} 
\end{figure}

In the penalty function, $c_x$ and $c_y$ indicate the minimum clearance between two interacting agents, representing the point of maximum penalty and serving as the ellipse's center. On the other hand, $r_x$ and $r_y$ define the desired clearance, determining the ellipse's elongation along the respective axes. When the distance from the center surpasses the radius ($r_x$ or $r_y$), the penalty rapidly diminishes, reflecting a negligible level of risk $\epsilon$, as shown in Fig.~\ref{fig:risk_field_1d}. Additionally, $P_x$ and $P_y$ modulate the relative contributions of the longitudinal and lateral directions to the penalty, while $P$ governs the overall extent of the ellipse. 

The values of $P_x$ and $P_y$ are adjusted based on the interaction type. For two vehicles driving in the same direction, longitudinal safety is prioritized, resulting in \( P_x = P_{\max} \) being higher than \( P_y = P_{\min} \). In contrast, vehicles approaching each other longitudinally should maintain greater lateral clearance to avoid collisions, leading to \( P_x = P_{\min} \) and \( P_y = P_{\max} \). The same configuration is applied to interactions involving static obstacles, as lateral clearance remains the primary concern. For intersection scenarios, where maintaining clearance in both directions is equally critical, the parameters are set such that \( P_x = P_y = P_{\max} \). 

In defining both geometric and dynamic risk penalties, we establish the ellipse center by adding the effective dimensions of the RL agent and the other interacting agent in a specific direction, as outlined in Eq.~\ref{eq:safe_distance}. Here, $L$ and $W$ represent a vehicle's length and width. In an intersection, we represent the vehicle's geometry with a circumcircle, ensuring enhanced lateral safety. These values are chosen because if the distances between two interacting agents in both directions are less than these values, a collision is inevitable and should result in substantial penalties.
\begin{equation}
c_{x} = \dfrac{ L_{agent} + L_{other}}{2}\; , 
\; c_{y} = \dfrac{W_{agent} + W_{other}}{2}
\label{eq:safe_distance}
\end{equation}
The key difference in calculating risk penalties lies in the desired ellipse radii. For the \textbf{geometric risk penalty} ${\mathcal{P}}_{\,risk}^{\,geom}$, the radii ($r_{x},r_{y}$) are set at predefined values, representing typical driver safety clearances. This ensures the agent maintains a consistent safety clearance, regardless of interaction dynamics, with these radii and proposed centers (Eq.~\ref{eq:safe_distance}) applied in Eq.~\ref{eq:risk_field}. The \textbf{dynamic risk penalty} ${\mathcal{P}}_{\,risk}^{\,dyn}\,$ adapts safety clearances by incorporating the velocities and accelerations of the interacting participants, ensuring the clearances are dynamically suited to the driving context. For interactions involving vehicles traveling in the same or opposite directions, as well as encounters with static obstacles, we develop a risk field based on worst-case analysis by adapting and refining the RSS framework. While RSS does account for lateral driving, its scope is limited to scenarios where vehicles approach each other. In this work, we extend the framework to include a wider range of lateral interactions.

The desired \textbf{dynamic longitudinal clearance} (ellipse longitudinal radius) is designed to ensure that both interacting agents can stop safely without colliding in the longitudinal direction. This calculation considers three possible scenarios, one of which involves the RL agent driving in the same direction as a leading vehicle, as demonstrated in Fig.~\ref{fig:long_same}. To account for the worst case, the safety clearance considers the leading vehicle suddenly braking at maximum deceleration ($a_{\max}^{\mathrm{brk}}$), while the RL agent accelerates toward it at its maximum acceleration ($a_{\max }^{\mathrm{acc}}$) for a reaction time ($\rho$), Eq.~\ref{eq:rho_velocity}, before applying its own minimum braking deceleration ($a_{\min }^{\mathrm{brk}}$), Eq.~\ref{eq:long_stop}. The desired safety clearance, represented as the ellipse radius in the longitudinal direction ($r_x^{\mathrm{leading}}$), is calculated as the difference between the RL agent’s stopping distance, including the acceleration and braking phases, and the leading agent’s stopping distance. This relationship is formalized in Eq.~\ref{eq:dyn_x_same}, where $v$ represents velocity and $d$ represents distance. For interactions involving static obstacles, the safety clearance is determined solely by the RL agent's stopping distance, calculated by Eq.~\ref{eq:dyn_x_same}, where ${v}_{\text{other}} = 0$.
\begin{align}
    d^{\,acc} &= v \cdot \rho+\frac{1}{2} \cdot a_{\max }^{\mathrm{acc}} \cdot \rho^2
    \label{eq:rho_velocity}\\
    d^{\,stop} &= \dfrac{\left(v+\rho \cdot a_{\max }^{\mathrm{acc}}\right)^2}{2 \, a_{\min }^{\mathrm{brk}}}
    \label{eq:long_stop}\\
    r_{x}^{leading} &= \left (d^{\,acc}_{agent} + d^{\,stop}_{agent}\right) - \dfrac{{v}_{other}^{2}}{2 \, a_{\max }^{\mathrm{brk}}}
    \label{eq:dyn_x_same}
\end{align}
In other scenarios,
where both agents are moving toward each other, as shown in Fig.~\ref{fig:long_opp}, the desired safety clearance accounts for the worst-case scenario where both agents accelerate towards each other with maximum acceleration $a_{\max }^{\mathrm{acc}}$ for a reaction time $\rho$ before starting to stop with minimum deceleration $a_{\min }^{\mathrm{brk}}$ as demonstrated in equation \ref{eq:dyn_x_opp}. 
\begin{equation}
    r_{x}^{approach} = \left (d^{\,acc}_{agent} + d^{\,stop}_{agent} \right ) + \left (d^{\,acc}_{other} + d^{\,stop}_{other} \right )
    \label{eq:dyn_x_opp}
\end{equation}
The desired \textbf{dynamic lateral clearance} (ellipse lateral radius) considers four possible motion scenarios between two agents. Previously examined cases, where agents either approach each other or the RL agent moves toward another agent moving away, also apply laterally ($r_{y}^{approach}$, $r_{y}^{leading}$), using lateral velocities, accelerations, and decelerations. Another case occurs when the RL agent moves away from the other agent while the latter approaches. If the RL agent's velocity is lower than the maximum velocity the other agent can achieve by accelerating for time $\rho$, as shown in Eq. \ref{eq:lat_safe}, a safety clearance must be maintained. This clearance is at least the difference between the distance the approaching agent can travel toward the RL agent and the distance the RL agent can move away, as described in Eq.~\ref{eq:dyn_y}. In the final case, where both agents are moving away from each other, no dynamic safety clearance is required. 
\begin{align}
\centering
    v_{other,\rho} &= v_{other} + \rho\,a_{\max }^{\mathrm{acc}}
    \label{eq:lat_safe} \\
    r_{y}^{away} &=
    \begin{cases}
        d^{\,acc}_{other} -v_{agent} \cdot \rho &, \text{If} \;  v_{ego} \leq  v_{other,\rho} \\ 
        0& , \text{otherwise}\\ 
    \end{cases}
    \label{eq:dyn_y}
\end{align}
For \textbf{intersecting} vehicles, as shown in Fig.~\ref{fig:interaction_types}, we use TTC instead of the ellipsoid function to calculate the dynamic risk penalty. TTC considers the positions $(x, y)$ and velocities $(v_x, v_y)$ of both agents, assuming a worst-case scenario where both accelerate toward each other at maximum acceleration. Unlike prior approaches~\cite{r20,li2018urban}, which assumed one-dimensional motion, our method is better suited for intersection scenarios with variations in both longitudinal and lateral directions. Specifically, we apply the circle algorithm~\cite{hou2015new}, which calculates the collision time $t_{ij}$ between the circumcircles of two vehicles $(i, j)$ by solving a quadratic Eq.~\ref{eq:ttc}, where $R_i$ and $R_j$ are the circumradii of the vehicles.
\begin{equation}
    \begin{bmatrix}
        \left (v_{xi} - v_{xj}\right )^2 + \left (v_{yi} - v_{yj}\right )^2\\ 
        \left (x_{i} - x_{j}\right ) \left (v_{xi} - v_{xj}\right ) + \left (y_{i} - y_{j}\right ) \left (v_{yi} - v_{yj}\right )\\ 
        \left (x_{i} - x_{j}\right )^2 + \left (y_{i} - y_{j}\right )^2 - (R_i + R_j)^2 
    \end{bmatrix}^T    
    \begin{bmatrix}
        {t_{\,ij}}^2 \\ 2\, t_{ij} \\ 1 
    \end{bmatrix}=0
    \label{eq:ttc}
\end{equation}
The estimated TTC is converted to a logarithmic scale to represent a risk measure, following the approach in~\cite{r7}. As shown in Eq.~\ref{eq:norm_ttc}, the risk field formulation uses $TTC_{\,\text{max}}$ as a threshold, where interactions exceeding this value are deemed non-risky and excluded from risk calculations. It should be noted that this term is normalized within \([0, 1]\), facilitating the assignment of a weight proportional to its importance in the risk-reward calculation.
\begin{equation}
{\mathcal{P}}_{\,risk}^{\,dyn} = -\log \left [\max\left ( 0.1,\min\left (\frac{TTC}{TTC_{\,\text{max}}},\; 1\right )\right ) \right ]
    \label{eq:norm_ttc}
\end{equation}
The overall risk awareness reward is a weighted sum of geometric and dynamic risk penalties, as illustrated in Eq.~\ref{eq:safety_reward}. The values of $w_{L_1^*}^{geom}$ and $w_{L_1^*}^{dyn}$ dictate the influence of each type of risk penalty on the overall reward. Since the RL agent may interact with \textbf{multiple agents} simultaneously, we select the one with the highest risk to contribute to the risk reward.
\begin{equation}
{\mathcal{R}}_{\,L_1^*} = -w_{L_1^*}^{geom}\cdot {\mathcal{P}}_{\,risk}^{\,geom} - w_{L_1^*}^{dyn} \cdot  {\mathcal{P}}_{\,risk}^{\,dyn}
\label{eq:safety_reward}
\end{equation}

\subsubsection*{\textbf{Progress (\(L_1\))}}
The progress objective focuses on maximizing the agent's efficiency in advancing along a reference route toward the destination, as illustrated in Eq.~\ref{eq:distance_reward}. The primary component of the objective is the distance covered by the agent in the direction of the goal during each step. To ensure consistency and transparency, the distance $d_{\,\text{traveled}}$ is normalized by the maximum distance the agent could theoretically traverse in a single step. This theoretical maximum is calculated as the product of the agent's maximum allowable velocity $v_{max}$ and the elapsed time between consecutive steps $\Delta t$. 
    \begin{equation}
        {\mathcal{R}}_{\,L_1} =  \dfrac{d_{\,\text{traveled}}}{v_{max} \cdot \Delta t}
        \label{eq:distance_reward}
    \end{equation}

\subsubsection*{\textbf{Driving style (\(L_2\))}} This level serves as a secondary consideration to shape the agent's progress along the route while adhering to desirable operational standards. The first term in Eq.~\ref{eq:progress_reward} encourages the agent to maintain a speed close to a predefined desired speed $v_{\text{\,desired}}$. This term is designed to shape a policy where progress toward the goal is achieved efficiently without excessively high or unusually low speeds. The second term emphasizes maintaining the lane, reflecting a standard practice in safe and predictable driving. Encouraging the agent to stay near the centerline this term supports lane discipline and minimizes erratic behavior.
    \begin{equation}
        {\mathcal{R}}_{\,L_2} = -w_{L_2}^{vel}\cdot \dfrac{\left| v_t - v_{\text{\,desired}}\right|}{v_{\text{\,desired}}}
        - w_{L_2}^{lane} \cdot \frac{\left| {\text{offset}}_{\,t}\right|}{\text{lane width}}
        \label{eq:progress_reward}
    \end{equation}

\subsubsection*{\textbf{Comfort (\(L_3\))}}
In autonomous driving, emphasis on comfort is imperative, given its profound impact on passenger acceptance. In the literature, two viewpoints on comfort are prevalent. One viewpoint associates comfort with the perceived risk of driving, where more risky driving leads to increased discomfort and agitation of passengers. For instance, headway has been incorporated into comfort calculations~\cite{r7}. Despite the validity of this definition, we omit it from our formulation of comfort since the perceived risk is extensively addressed in our risk-awareness objective, rendering duplicate terms unnecessary.

The other viewpoint focuses on comfort concerning the dynamic aspects of vehicle motion~\cite{bellemcomfort2018}. The comfort reward, illustrated in Eq.~\ref{eq:comfort_reward}, is the sum of the ratios of steering rate $\dot{\theta}_{t}$, acceleration $a_t$, and jerk $j_t$ to their respective maximum values. Acceleration is constrained between \([-8, 8]\) m/s\(^2\), in accordance with the statistical limits proposed in~\cite{girase2021physically}. The steering rate is penalized to encourage smoother steering. The maximum steering rate is equal to the product of the current velocity and maximum curvature, with curvature confined between \([-0.3, 0.3]\) m\(^{-1}\) similar to~\cite{girase2021physically}. Jerk, the derivative of acceleration, is associated with strong vibrations, particularly at high jerk values. The comfort term penalizes jerk to promote smoother driving with reduced vibrations. The maximum jerk equals the maximum acceleration divided by the time between steps.
\begin{equation}
{\mathcal{R}}_{\,L_3} = \dfrac{-1}{3} \cdot\left (
\dfrac{a_t}{a_{max}}+ \dfrac{\dot{\theta}_{t}}{v_{t}\cdot \kappa_{max}}+
\dfrac{j_t}{a_{max}/ \Delta t}
\right ) 
\label{eq:comfort_reward}
\end{equation}
\section{Experimental Setup}
\label{sec:experiments}
The section discusses the experimental setup employed in this work. First, we explain the employed RL agent, covering its observation space, action space, and the learning algorithm. Additionally, we elaborate on the driving scenario used for evaluating the agent's performance.
\begin{table*}[]
\renewcommand{\arraystretch}{1.3}
\caption{The evaluation metrics assessed on agents trained with different combinations of the proposed reward function levels, using a hold-out set of intersection environments. These environments were randomly generated with traffic density varying from \textit{0.5} to \textit{1.0}.}
\centering
\resizebox{\textwidth}{!}{
\begin{tabular}{lrrr|rrr|rrr}
\toprule
& \multicolumn{3}{c|}{Traffic Density 0.5}                       & \multicolumn{3}{c|}{Traffic Density 0.75}                         & \multicolumn{3}{c}{Traffic Density 1.0}                           \\
Metrics                                        & $L_{0-1}$ & $L_{0-3}$ & $L_{complete}$ & $L_{0-1}$  & $L_{0-3}$ & $L_{complete}$  &
$L_{0-1}$        & $L_{0-3}$ & $L_{complete}$    \\ \midrule
Success (\%) $\boldsymbol{\uparrow}$           & 34.2              & 64.0             & \textbf{73.1}           & 26.6            & 45.5                   & \textbf{60.6}          & 21.2           & 36.3                   & \textbf{48.4}           \\
Off-road (\%) $\boldsymbol{\downarrow}$        & 23.2              & \textbf{0.2}     & 7.3                     & 17.7            & \textbf{0.1}           & 8.9                    & 16.7           & \textbf{0.0}           & 12.8                    \\
Collision (\%) $\boldsymbol{\downarrow}$       & 42.5              & 35.5             & \textbf{19.6}           & 55.6            & 53.9                   & \textbf{30.5}          & 61.9           & 62.7                   & \textbf{38.8}           \\
Time-out (\%) $\boldsymbol{\downarrow}$        & 0.1               & 0.3              & \textbf{0.0}            & 0.1             & 0.5                    & \textbf{0.0}           & 0.2            & 1.0                    & \textbf{0.0}            \\ \midrule
Cumulative Reward $\boldsymbol{\uparrow}$      & $-1.07 \pm 2.86 $ & $0.21\pm1.98$    & $\mathbf{0.78}\pm 1.53$ & $-1.38\pm2.83$ & $-0.49\pm2.23$         & $\mathbf{0.32}\pm1.69$ & $-1.73\pm2.91$ & $-1.04\pm2.43$         & $\mathbf{-0.10}\pm1.82$ \\ \midrule
Route Progress $\boldsymbol{\uparrow}$         & $0.57 \pm0.34$    & $0.72\pm0.30$    & $\mathbf{0.79}\pm0.25$  & $0.49\pm0.34$   & $0.60\pm0.34$          & $\mathbf{0.71}\pm0.30$ & $0.43\pm0.34$  & $0.52\pm0.35$          & $\mathbf{0.63}\pm 0.32$ \\
Average Velocity (m/s) $\boldsymbol{\uparrow}$ & $2.75\pm0.77$     & $3.26\pm0.67$    & $\mathbf{3.47}\pm0.67$  & $2.59\pm0.78$   & $\mathbf{3.13}\pm0.74$ & $3.11\pm0.86$          & $2.42\pm0.81$  & $\mathbf{2.91}\pm0.86$ & $2.78\pm0.95$           \\ \bottomrule
\end{tabular}}
\label{table:final_metrics}
\vspace{-0.2cm}
\end{table*}

\subsection{RL Agent Description}
In this work, we employ a multimodal perception observation space consisting of a frontal RGB camera with a $128 \times 128$ resolution and a LiDAR point cloud discretized into a $128 \times 128$ grid map with two height bins. 
Unlike previous studies~\cite{xiao2020multimodal,chen2021interpretable}, we refrain from encoding reference route or obstacle level information into additional image channels. 
Instead, we condition our network learning on discrete navigational commands (follow lane, turn left/right), similar to~\cite{chitta2022transfuser}. Additionally, vehicle measurements, including longitudinal and angular velocities and longitudinal and lateral accelerations, are incorporated into the observation space. 
We adopted TransFuser~\cite{chitta2022transfuser} as a backbone to encode our input observations into a latent space. TransFuser is an efficient transformer-based architecture that fuses features from RGB images and LiDAR grid maps at multiple scales using cross-attention, thereby enhancing spatial awareness. 

This work utilizes an RL agent for vehicle trajectory planning, leveraging a Frenet trajectory planner~\cite{bogdoll2024informed} for its adaptability and road-aware trajectory generation. Based on the current observation, the agent estimates two boundary conditions $(v_f, d_f)$ as a discrete output action. $v_f$ represents the desired velocity at the end of the planning horizon, while $d_f$ denotes the desired lateral offset from the lane centerline. Subsequently, a Frenet planner generates a trajectory fulfilling these boundary conditions, and a simple PID controller is used to track the planned trajectory. The RL agent is trained using DQN~\cite{mnih2013playing}, a well-established deep RL algorithm commonly employed in autonomous driving tasks for its effectiveness in discrete action spaces.
\subsection{Traffic Scenarios} 
We focus on driving scenarios where the agent begins at a predefined distance from an unsignalized intersection, requiring it to approach and safely navigate through the intersection to reach its goal. This emphasis is motivated by the prevalence of unsignalized intersections in urban environments, where implicit coordination with other vehicles is essential, and safety-critical situations frequently arise~\cite{khaitan2022state}. Although our framework is applicable to various driving contexts, intersections offer a unique set of challenges that make them an ideal testbed for our risk field reward. The traffic scenarios are implemented in Town04 of Carla~\cite{dosovitskiy2017carla}. During training, vehicles and static obstacles are randomly placed along three T-intersections and four 4-way intersections. The attributes of the added vehicles, including shape, velocity, and lane offset, are randomized to enhance the agent’s robustness and generalization. The agent’s performance is evaluated on a hold-out set comprising one T-intersection and two 4-way intersections that were not encountered during training.
\subsection{Baselines and Evaluation Metrics} 
We evaluate the effectiveness of our reward function by analyzing the impact of incrementally adding levels from the proposed reward hierarchy. The first variant involves training an agent using a reward that combines traffic rule compliance ($L_{0}$) and progress ($L_{1}$), denoted as $L_{0-1}$. The second variant expands the reward by including two additional progress-shaping objectives: driving style ($L_{2}$), which promotes lane-keeping and maintaining a desired velocity, and comfort ($L_{3}$), referred to as $L_{0-3}$. The final variant $L_{\text{complete}}$ incorporates a risk-awareness objective to better balance the agent’s progress and minimize risks and collisions. 

To maintain consistency in evaluation, all agents are trained for the same number of steps, utilizing identical architectures, learning hyperparameters, and reward parameters. The parameters used in the proposed reward function are provided in Table~\ref{table:parameters} and were selected to closely align with those used in RSS~\cite{rss}. The performance of the trained agents is assessed on a separate set of intersections with varying traffic densities, where traffic density is defined as the proportion of actors (NPCs and obstacles) present relative to the maximum allowable number in the scenario. The evaluation metrics include cumulative rewards per episode and a range of standard driving performance metrics. These metrics cover terminal statistics such as the percentage of successful episodes, off-road deviations, collisions, and behavioral metrics like route progress and average velocity.
\begin{table}[]
\renewcommand{\arraystretch}{1.1}
\caption{The parameters of the proposed reward function.}
\centering
\resizebox{0.96\columnwidth}{!}{
\begin{tabular}{c c | c c | c c}
\toprule
\textbf{Parameter} & \textbf{Value} & \textbf{Parameter} & \textbf{Value} & \textbf{Parameter} & \textbf{Value} \\ 
\midrule
$\beta$            & 0.25          & $P$                  & 4              & $\rho$             & 0.3            \\
$v_{\max}$         & 6.0           & $a^{\text{acc}}_{\max, x}$ & 6.0      & TTC$_{\max}$       & 7.0            \\
$w_{\text{Terminal}}$  & 50.0     & $a^{\text{acc}}_{\max, y}$ & 0.2      & $w^{\text{geom}}_{L^*_1}$ & 0.5 \\
$r_x$             & 2.0           & $a^{\text{brk}}_{\min, x}$ & 4.0      & $w^{\text{dyn}}_{L^*_1}$  & 0.5 \\
$r_y$             & 0.5           & $a^{\text{brk}}_{\max, x}$ & 8.0      & $v_{\text{des}}$   & 4.0            \\
$P_{\min}$        & 2.0           & $a^{\text{brk}}_{\min, y}$ & 0.4      & $w^{\text{vel}}_{L_2}$    & 0.5 \\
$P_{\max}$        & 4.0           & $a^{\text{brk}}_{\max, y}$ & 0.8      & $w^{\text{lane}}_{L_2}$   & 0.5 \\
\bottomrule
\end{tabular}
}
\label{table:parameters}
\vspace{-0.3cm}
\end{table}

\section{Evaluation}
\label{sec:evaluation}

As shown in Table \ref{table:final_metrics}, the evaluation results demonstrate that incorporating risk-awareness into the reward function significantly enhances RL agent performance across different traffic densities. Safety metrics highlight the benefits of \( L_{\text{complete}} \). At a density of 0.5, the collision rate decreases from 42.5\% with \( L_{0-1} \) to 35.5\% with \( L_{0-3} \), reaching a minimum of 19.6\% with \( L_{\text{complete}} \). At a density of 1.0, where traffic interactions become more complex, \( L_{\text{complete}} \) achieves the lowest collision rate of 38.8\%, a clear improvement over \( L_{0-3} \) (62.7\%) and \( L_{0-1} \) (61.9\%). Off-road violations follow a similar trend. At lower density, the rate is highest with \( L_{0-1} \) at 23.2\% and drops to 0.2\% with \( L_{0-3} \), while \( L_{\text{complete}} \) balances safety and lane keeping with a rate of 7.3\%. In high-density conditions, \( L_{\text{complete}} \) show a slight increase in off-road incidents to 12.8\%, reflecting a necessary trade-off between maintaining safe distances and ensuring task completion.

The cumulative reward metric further highlights the overall effectiveness of each reward function. At a density of 0.5, \( L_{\text{complete}} \) achieves the highest reward of 0.78, improving upon \( L_{0-3} \) (0.21) and \( L_{0-1} \) (-1.07). This advantage remains evident at a density of 1.0, where \( L_{\text{complete}} \) maintains the best reward at -0.10, while \( L_{0-3} \) and \( L_{0-1} \) drop to -1.04 and -1.73, respectively. A similar trend is observed in route progress, where \( L_{\text{complete}} \) consistently outperforms other models, reaching 0.79 at low density and 0.63 in high-density conditions. The average driving velocity indicates that \( L_{\text{complete}} \) enables agents to maintain efficient speeds while reducing unnecessary braking. At a density of 0.5, it reaches 3.47 m/s, surpassing \( L_{0-3} \) (3.26 m/s) and \( L_{0-1} \) (2.75 m/s). Even in dense traffic, \( L_{\text{complete}} \) maintains a competitive velocity of 2.78 m/s, close to \( L_{0-3} \) (2.91 m/s) while remaining higher than \( L_{0-1} \) (2.42 m/s).  

These results confirm that risk-aware rewards lead to safer driving behavior. The proposed \( L_{\text{complete}} \) reward function achieves the highest success rates, reduces collisions, and maintains efficient route progress across different traffic densities. While progress-only rewards struggle with high accident rates and intermediate rewards remain suboptimal in dense traffic. These findings emphasize the importance of designing reward functions that balance risk and efficiency, making them suitable for real-world driving applications.

\section{Conclusion}
\label{sec:conclusion}
This paper introduces a hierarchical reward function to enhance RL for autonomous driving, providing a structured approach to balancing safety, progress, comfort, and rule conformance. The reward function employs normalized objectives and a systematic weight-setting methodology, ensuring transparency and scalability in managing the conflicting goals inherent in driving tasks. A key contribution is a risk-aware objective, which incorporates worst-case analysis and a two-dimensional ellipsoid function to model geometric and dynamic aspects of interactions, enabling agents to effectively anticipate and mitigate risks. The work further details how ellipse parameters are adapted to various interaction types common in driving scenarios. The impact of the proposed reward function is assessed by training DQN agents on different subsets of the reward structure and evaluated in diverse unsignalized intersection scenarios. Results demonstrate that the driving risk objective significantly enhances the agent’s risk awareness and driving behavior while maintaining efficient progress, showcasing its potential for advancing safe and reliable driving agents.
{
    \bibliographystyle{IEEEtran}
    \bibliography{references}
}

\end{document}